\pdfoutput=1
\documentclass[11pt]{article}

\usepackage[]{ACL2023}

\usepackage{times}
\usepackage{latexsym}
\usepackage[T1]{fontenc}
\usepackage[utf8]{inputenc}

\usepackage{microtype}
\usepackage{pifont}

\usepackage{inconsolata}

\usepackage{graphicx}
\usepackage{amsmath}
\usepackage{amsfonts}
\usepackage{amsthm}
\usepackage{multirow}
\usepackage{array}
\usepackage{color}
\usepackage{makecell}
\usepackage{etoolbox}
\usepackage{url}

\graphicspath{ {images/} }

\title{Opinion Tree Parsing for Aspect-based Sentiment Analysis}

\author{Xiaoyi Bao$^{1}$,  Xiaotong Jiang$^{1}$, Zhongqing Wang$^{1}$\thanks{\quad Corresponding author},  Yue Zhang$^{2,3}$, and Guodong Zhou$^{1}$ \\
  $^{1}$Natural Language Processing Lab, Soochow University, Suzhou, China \\
  $^{2}$School of Engineering, Westlake University\\
  $^{3}$Institute of Advanced Technology, Westlake Institute for Advanced Study\\
  \texttt{ p2213545413@outlook.com,devjiang@outlook.com}\\
  \texttt{yue.zhang@wias.org.cn}\\
  \texttt{\{wangzq,gdzhou\}@suda.edu.cn}\\
   }

\begin{document}

\maketitle

\begin{abstract}

Extracting sentiment elements using pre-trained generative models has recently led to large improvements in aspect-based sentiment analysis benchmarks.
However, these models always need large-scale computing resources, and they also ignore explicit modeling of structure between sentiment elements.
To address these challenges, we propose an opinion tree parsing model, aiming to parse all the sentiment elements from an opinion tree, which is much faster, and can explicitly reveal a more comprehensive and complete aspect-level sentiment structure.
In particular, we first introduce a novel context-free opinion grammar to normalize the opinion tree structure. 
We then employ a neural chart-based opinion tree parser to fully explore the correlations among sentiment elements and parse them into an opinion tree structure.
Extensive experiments show the superiority of our proposed model and the capacity of the opinion tree parser with the proposed context-free opinion grammar.
More importantly, the results also prove that our model is much faster than previous models. Our code can be found in \url{https://github.com/HoraceXIaoyiBao/OTP4ABSA-ACL2023}.
\end{abstract}

\section{Introduction}

Aspect-based sentiment analysis (ABSA) has drawn increasing attention in the community, which includes four subtasks: aspect term extraction, opinion term extraction, aspect term category classification and aspect-level sentiment classification. 
The first two subtasks aim to extract the aspect term and the opinion term appearing in one sentence.
The goals of the remaining two subtasks are to detect the category and sentiment polarity towards the extracted aspect term.

\begin{figure}[t]
	\centering
	\includegraphics[width=220px]{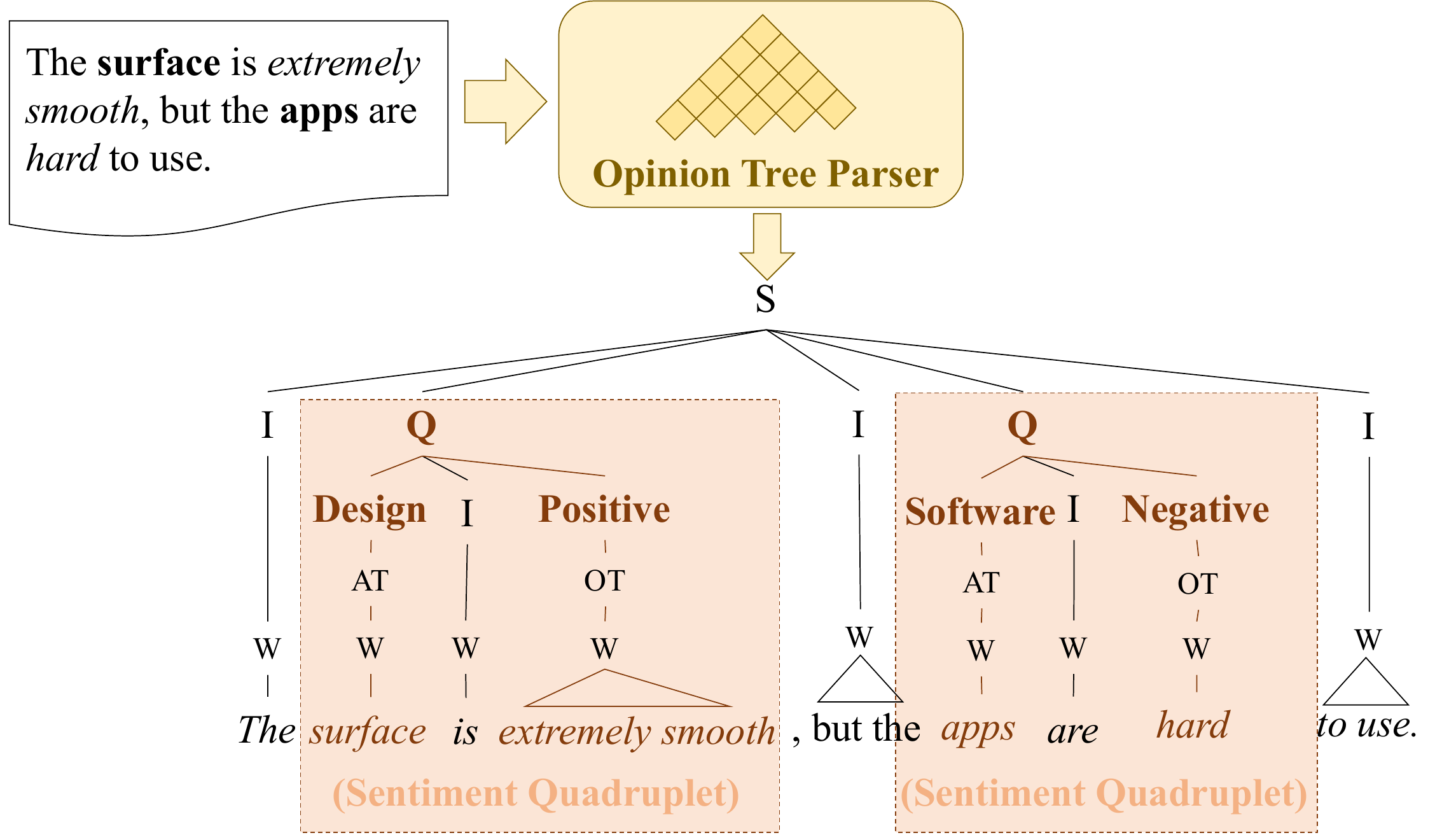}
	\caption{\label{figure-example0} Example of opinion tree parsing.}
\end{figure}

Previously, most ABSA tasks are formulated as either sequence-level~\cite{QiuLBC11,PengXBHLS20,CaiXY20} or token-level classification problems~\cite{TangQFL16}. 
However, these methods usually suffer severely from error propagation because the overall prediction performance hinges on the accuracy of every step~\cite{PengXBHLS20}. 
Therefore, recent studies tackle the ABSA problem with a unified generative approach. 
For example, they treat the class index~\cite{YanDJQ020} or the desired sentiment element sequence~\cite{Zhang0DBL20,ZhangD0YBL21} as the target of generation model. 
More recently, \citet{BaoWJXL22} addresses the importance of correlations among sentiment elements (e.g., aspect term, opinion term), and proposes an opinion tree generation model, which aims to jointly detect all sentiment elements in a tree structure. 

The major weakness of generative approaches is
the training and inference efficiency, they always need large-scale computing resources.
In addition, these generative approaches lack certain desirable properties. 
There are no structural guarantees of structure well-formedness, i.e. the model may predict strings that can not be decoded into valid opinion trees, and post-processing is required. 
Furthermore, predicting linearizations ignores the implicit alignments among sentiment elements, which provide a strong inductive bias.

As shown in Figure~\ref{figure-example0}, we convert all the sentiment elements into an opinion tree and design a neural chart-based \emph{opinion tree parser} to address these shortcomings.
The opinion tree parser is much simpler and faster than generative models.
It scores each span independently and performs a global search over all possible trees to find the highest-score opinion tree~\cite{KleinK18,KitaevCK19}.
It explicitly models tree structural constraints through span-based searching and yield alignments by construction, thus guaranteeing tree structure well-formedness.

One challenge to the above is that not all the review texts contain standard sentiment quadruplets (i.e., aspect term, opinion term, aspect category, and polarity) which can be easily formed in an opinion tree~\cite{BaoWJXL22}. For example, there may be more than one opinion term correlated with an aspect term and vice versa. In addition, aspect or opinion terms might be implicit.
According to our statistics, such irregular situations appear in more than half of review texts. 
In this study, we propose a novel \emph{context-free opinion grammar} to tackle these challenges.
The grammar is generalized and well-designed, it is used to normalize the sentiment elements into a comprehensive and complete opinion tree.
Furthermore, it contains four kinds of conditional rules, i.e.,  one-to-many, mono-implicit, bi-implicit, cross-mapping, which are used to solve the irregular situations in opinion tree parsing.

The detailed evaluation shows that our model significantly advances the state-of-the-art performance on several benchmark datasets. 
In addition, the empirical studies also indicate that the proposed opinion tree parser with context-free opinion grammar is more effective in capturing the sentiment structure than generative models. More importantly, our model is much faster than previous models.  

	

\section{Related Work}



As a complex and challenging task, aspect-based sentiment analysis (ABSA) consists of numerous sub-tasks. The researches on ABSA generally follow a route from handling single sub-task to complex compositions of them.
The fundamental sub-tasks focus on the prediction of a single sentiment element, such as extracting the aspect term~\cite{QiuLBC11,TangQFL16,wang-etal-2021-progressive}, 
detecting the mentioned aspect category~\cite{BuRZYWZW21,HuZZCSCS19}, 
and predicting the sentiment polarity for a given aspect~\cite{TangQFL16,chen-etal-2022-discrete,liu-etal-2021-solving,seoh-etal-2021-open,zhang-etal-2022-ssegcn}.

Since the sentiment elements are natural correlated, many studies focus on exploring the joint extraction of pairwise sentiment elements, including aspect and opinion term extraction~\cite{XuLLB20,li-etal-2022-generative};
aspect term extraction and its  polarity detection~\cite{ZhangQ20}; 
aspect category and  polarity detection~\cite{CaiTZYX20}.
Furthermore, recent studies also  employed end-to-end models to extract all the sentiment elements in triplet or quadruple format~\cite{PengXBHLS20,WanYDLQP20,CaiXY20,ZhangD0YBL21,chen-etal-2022-enhanced,mukherjee-etal-2021-paste}.

More recently, studies using pre-trained encoder-decoder language models show great improvements in ABSA~\cite{ZhangD0YBL21}.
They either treated the class index~\cite{YanDJQ020} or the desired sentiment element sequence~\cite{Zhang0DBL20} as the target of the generation model. 
in addition, \citet{BaoWJXL22} addressed the importance of correlations among sentiment elements, and proposed an opinion tree generation model, which aims to jointly detect all sentiment elements in a tree structure. However, the generative models always need large-scale computing resources, they also cannot guarantee the structure well-formedness, and ignores the implicit alignments among sentiment elements.



In this study, we propose a novel opinion tree parser, which aims to model and parse the sentiment elements from the opinion tree structure.
The proposed model shows significant advantages in both decoding efficiency and performance as it is much faster and more effective in capturing the sentiment structure than generative models.
Furthermore, we design a context-free opinion grammar to normalize the opinion tree structure, and improve parser's applicability decisions for complex compounding phenomena.






\section{Overview of Proposed Model}

\begin{figure*}[t]
	\centering
	\includegraphics[width=320px]{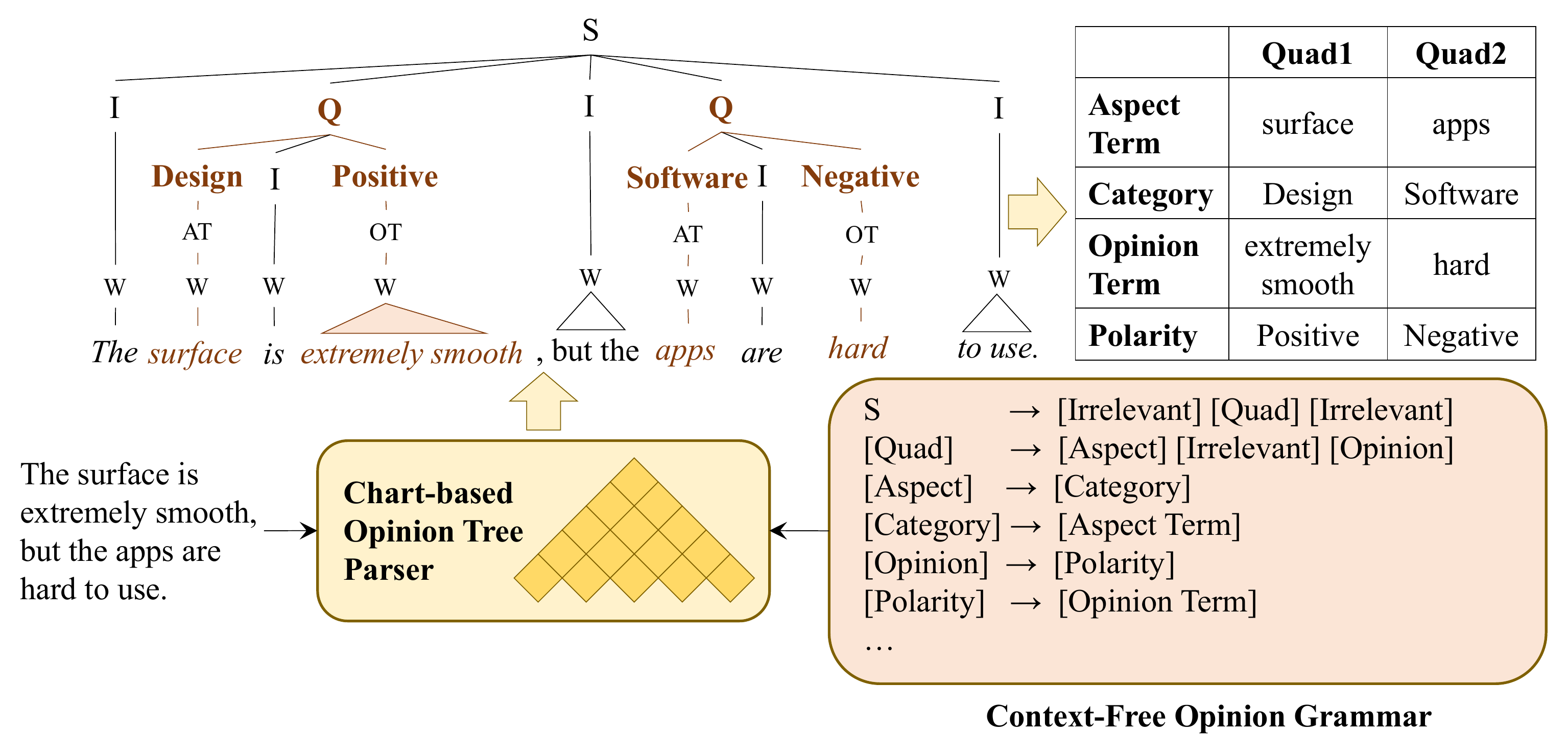}
	\caption{\label{figure-model} Overview of proposed model.}
\end{figure*}

Aspect-based sentiment analysis aims to extract all kinds of sentiment elements and their relations from review text.
Basically, there are four kinds of sentiment elements in the review text: \emph{aspect term} denotes an entity and its aspect indicating the opinion target, which is normally a word or phrase in the text; \emph{aspect category} represents a unique predefined category for the aspect in a particular domain; \emph{opinion term} refers the subjective statement on an aspect, which is normally a subjective word or phrase in the text; \emph{polarity} is the predefined semantic orientation (e.g., positive, negative, or neutral) toward the aspect.

As shown in Figure~\ref{figure-model}, we convert all the sentiment elements into an opinion tree, and we design a chart-based opinion tree parser with context-free opinion grammar to parse the opinion tree from review text.
In particular, we firstly propose a context-free opinion grammar to normalize the sentiment elements into an opinion tree.
We then perform a neural chart-based opinion tree parser to parse the opinion tree structure from a given review text.
Since all the sentiment elements are normalized into the opinion tree, it is easy to recover them from the tree.
In the next two sections, we will discuss the context-free opinion grammar and the opinion tree parser in detail.

\section{Context-Free Opinion Grammar}

In this study, we propose a novel context-free opinion grammar to normalize the opinion tree structure.
In the below of this section, we first introduce basic definitions of context-free opinion grammar. After that, we give some conditional rules to solve irregular situations and show some examples to illustrate the effectiveness of proposed grammar.

\subsection{Basic Definitions}

A context-free opinion grammar (CFOG) is a tuple $G = (N, \Sigma, P, S)$, where $N$ and $\Sigma$ are finite, disjoint sets of non-terminal and terminal symbols, respectively, Table~\ref{table-nonterminal} gives the notation of non-terminals.  $S \in N$ is the start symbol and $P$ is a finite set of rules.
Each rule has the form $ A \to \alpha$, where $A \in N$, $\alpha \in V_I^*$ and $V_I=N \cup \Sigma$. 

\begin{table}[!tp]
	\centering \small
	\begin{tabular}{c|l}
		\hline
		\bf Name & \bf Description \\ \hline
		Q &  Quad of sentiment elements  \\ \hline
		I &  Irrelevant content (\emph{e.g., the, but, are})   \\ \hline
		A & Aspect pair (\emph{Category, Aspect Term}) \\ \hline
		O & Opinion pair (\emph{Polarity, Opinion Term}) \\ \hline
		C & Category of aspect (\emph{e.g., Design, Software}) \\ \hline
		P & Polarity towards the aspect (\emph{Positive, Negative}) \\ \hline
		AT& Aspect term (\emph{e.g., surface, apps}) \\ \hline
		OT& Opinion term (\emph{e.g., smooth, hard}) \\ \hline
		W & Word \\ \hline
	\end{tabular}
	\caption{Notation of non-terminals.} \label{table-nonterminal}
\end{table}

\begin{figure*}[t]
	\centering
	\includegraphics[width=360px]{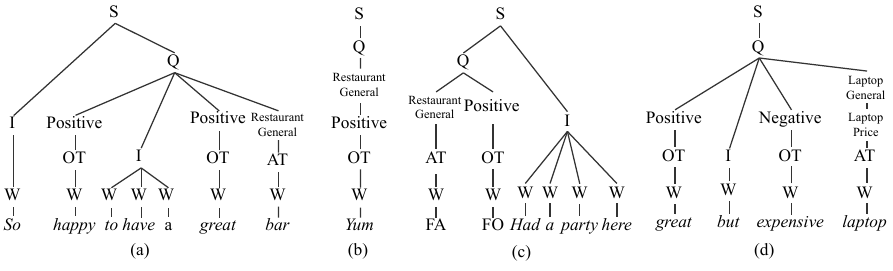}
	\caption{\label{conditional-rules} Examples of opinion trees with conditional rules and pruning approach.}
\end{figure*}

The top of Figure~\ref{figure-model} gives an example of opinion parsing tree.
Each terminal in the tree is either an irrelevant word or a sentiment element like aspect  or opinion term.
Each non-terminal combines terminals or non-terminals to create a sub-tree of sentiment elements.
In order to make the description as clear as possible, we begin with the basic rules allowed by our grammar: \newline

S $\to$ I Q I \quad \textcolor{gray}{// S $\to$ irrelevant content,quad,irrelevant content}

Q  $\to$ A I O | O I A | $\epsilon$ \quad \textcolor{gray}{// quad $\to$ (aspect,opinion) or (opinion,aspect)} 

Q  $\to$ Q I Q  \textcolor{gray}{// multiple quads}

A $\to$  C \quad \textcolor{gray}{// aspect $\to$ category}

C $\to$ AT \quad \textcolor{gray}{// category $\to$ aspect term}

O $\to$ P \quad \textcolor{gray}{// opinion $\to$ polarity}

P $\to$ OT \quad \textcolor{gray}{// polarity $\to$ opinion term}

AT $\to$ W \quad \textcolor{gray}{// aspect term $\to$ word}

OT $\to$ W \quad \textcolor{gray}{// opinion term $\to$ word}

I $\to$ W \quad\textcolor{gray}{// irrelevant content $\to$ word}

W$\to$  W W | $\epsilon$  

W $\to$   happy | to | great | party | but | have | ...

C $\Leftrightarrow$  Surface | Laptop | ... \quad \textcolor{gray}{// $C$ is replaced with a certain category} 

P $\Leftrightarrow$  Positive | Negative | Neutral \quad \textcolor{gray}{// $P$ is replaced with a certain polarity} 
\newline       

In the above notations, the rules bring out the grammatical relations among the elements of a standard sentiment quadruplet. 
For example, $I$ is used to define the irrelevant content in the review sentence, and $Q$ is used to describe a sentiment quadruple.
In addition, the components of quadruple, i.e., $A$ and $O$, are used to denote the aspect pair (category $C$ and aspect term $AT$) and opinion pair (polarity $P$ and opinion term $OT$). 
Since the opinion trees built under the above grammar may be too complicated, we adopt a  pruning approach to reduce the duplication in the trees, detail discussion of pruning can be found in Appendix~\ref{appendixA}.


\subsection{Conditional Rules}

Although the basic rules can be used to parse an opinion tree with standard quadruplets, they cannot handle irregular situations.
In this subsection, we introduce conditional rules to improve rule applicability  for complex compounding phenomena.

\textbf{One-to-Many} means that there is more than one opinion term correlated with an aspect term, and vice versa.
For example, in the review sentence ``So \emph{happy} to have a \emph{great} \emph{bar}'', both opinion terms ``\emph{happy}'' and ``\emph{great}'' are mapped to the same aspect term ``\emph{bar}''. In this study, we attach successor elements to the preceding  one and charge the rule $A$ and $O$ below for solving this situation:  \newline

\textcolor{gray}{// multiple aspects map to one opinion}

A $\to$  A I A    

\textcolor{gray}{// multiple opinions map to one aspect}

O $\to$ O I O  \newline

Then, the above cause can be correctly parsed through these two new rules. The example of parsing result is shown in Figure~\ref{conditional-rules}(a).

\textbf{Mono-Implicit} means that either aspect term or opinion term is missing in the review text. Given a review sentence ``\emph{Yum}'', only an opinion term appears in the sentence. 
For solving this problem, we attach the opinion to corresponding aspect node or attach the aspect to corresponding opinion node: \newline

\textcolor{gray}{// implicit aspect term  }

Q $\to$ C; C $\to$ O \quad \textcolor{gray}{// quad $\to$category$\to$opinion }

\textcolor{gray}{// implicit opinion term  }

Q $\to$ P; P $\to$ A \quad \textcolor{gray}{// quad $\to$polarity $\to$ aspect }\newline

An example of this solution can be found in Figure~\ref{conditional-rules}(b).

\textbf{Bi-Implicit} denotes that both the aspect term and opinion term are missing in the review text. As shown in the review sentence ``Had a party here'', although we know that the authors express a positive opinion,  both aspect term and opinion term do not appear in the sentence.
To solve the situation, we insert two fake tokens $FA$ and $FO$ at the beginning of a sentence as the fake aspect and opinion term.
Then, we can use standard rules to parse such sentences with implicit aspect and opinion. Figure~\ref{conditional-rules}(c) gives an example of this solution.

\textbf{Cross-Mapping} means that there are more than one aspect category and opinion polarity on the review text, and their correlations are many-to-many.
For example, in the review sentence ``\emph{Great} but \emph{expensive} laptop'', there are two categories ``Laptop General'' and ``Laptop Price'' towards the aspect term ``laptop''. Meanwhile, the opinions towards these two categories are different. The author feels ``great'' about the ``Laptop General'', but thinks the ``Laptop Price'' is ``expensive''.  The solution of such situation is shown in below: \newline

 \textcolor{gray}{// two categories and two opinion terms towards one aspect term }

A $\to$ C$_1$; C$_1$ $\to$ C$_2$; C$_2$ $\to$ AT  

 \textcolor{gray}{// two categories and two opinion terms towards one opinion term }

O $\to$ P$_1$; P$_1$ $\to$ P$_2$; P$_2$ $\to$ OT  \newline

Then, we use the shortest path to detect the correlation between aspect category and opinion term.
As shown in Figure~\ref{conditional-rules}(d), since the distance between ``Laptop General'' and ``great'' is shorter than  `` \emph{expensive} '', we connect ``Laptop General'' with `` \emph{great}'', and then connect ``Laptop Price'' with `` \emph{expensive} ''.

In summary, based on the basic and conditional rules, the proposed context-free opinion grammar can solve most situations in aspect-based sentiment analysis, and would help parse a comprehensive and complete opinion tree.

\section{Opinion Tree Parser}

\begin{figure}[t]
	\centering
	\includegraphics[width=130px]{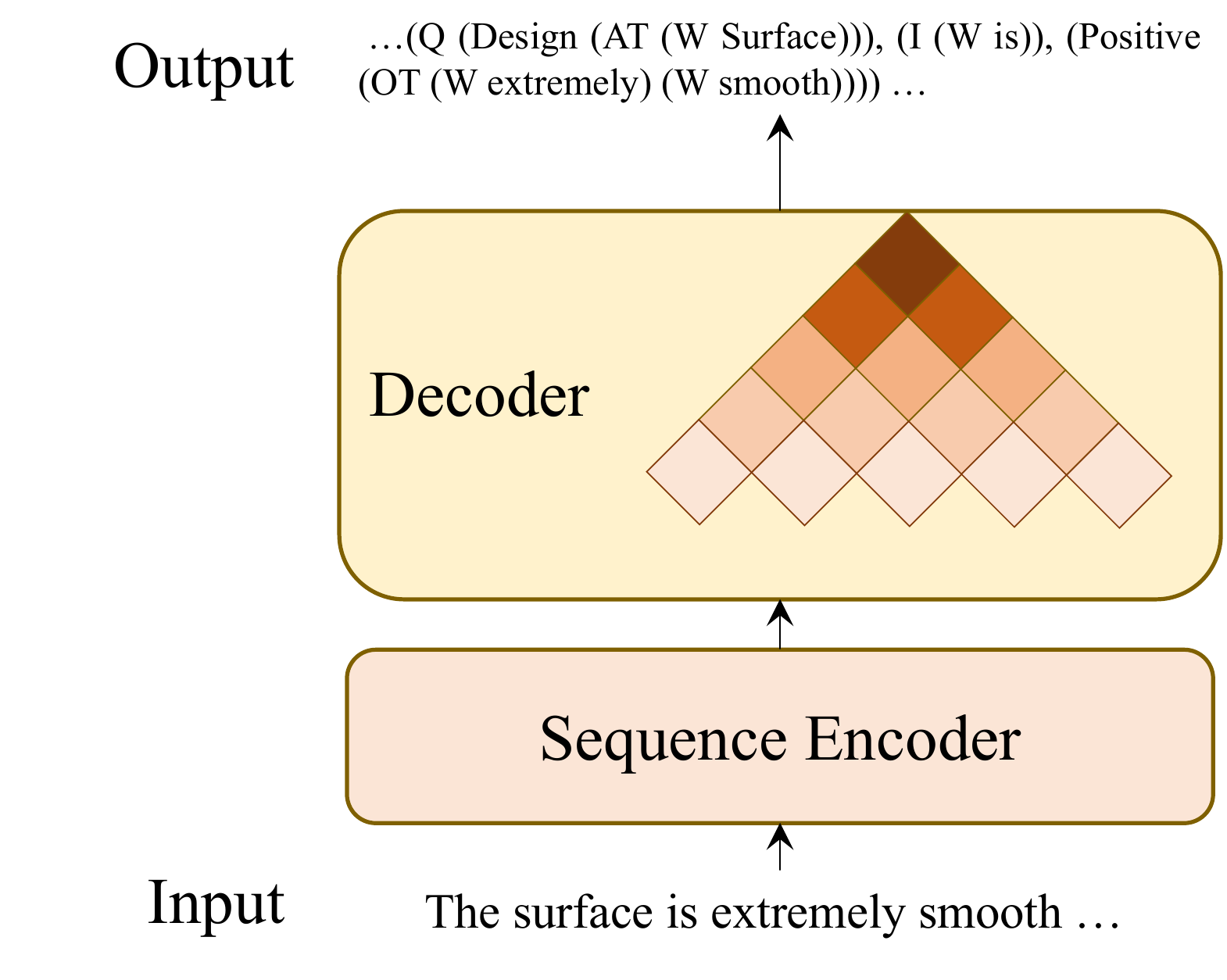}
	\caption{\label{figure-parser} Neural chart-based pinion tree parser.}
\end{figure}

In this study, we employ a neural chart-based opinion tree parser to parse sentiment elements from the opinion tree structure.
As shown in Figure~\ref{figure-parser}, the opinion tree parser follows an encoder-decoder architecture~\cite{KleinK18,KitaevCK19,CuiY022}. 
It scores each span independently and performs a global search over all possible trees to find the highest-score opinion tree.
In particular, the process of opinion tree parsing can be separated into two stages: context-aware encoding and chart-based decoding, we will discuss these in the below subsections.

\subsection{Span Scores and Context-Aware Encoding}

Given a review text $X = \{x_1, ..., x_n\}$, its corresponding opinion parse tree $T$ is composed by a set of labeled spans:
\begin{equation}
T=\{(i_t,j_t,l_t)\}|_{t=1}^{|T|}
\end{equation}
where $i_t$ and $j_t$ represent the $t$-th  span's fencepost positions and $l_t$ represents the span label. 

We use a self-attentive encoder as the scoring function $s(i, j)$, and a chart decoder to perform a global-optimal search over all possible trees to find the highest-scoring tree given the review text. 
In particular, given an input review text $X = \{x_1, ..., x_n\}$, a list of hidden representations $H^n_1 = \{h_1, h_2, ..., h_n\}$ is produced by the encoder, where $h_i$ is a hidden representation of the input token $x_i$. 
The representation of a span $(i, j)$ is constructed by:
\begin{equation}
v_{i,j}=h_j-h_i
\end{equation}

Finally, $v_{i,j}$ is fed into an MLP to produce real valued scores $s(i, j, )$ for all labels:
\begin{equation}
s(i,j)=W_2ReLU(W_1v_{i,j}+b_1)+b_2
\end{equation}
where $W_1$, $W_2$, $b_1$ and $b_2$ are trainable parameters, $W_2 \in R^{|H| \times |L|}$ can be considered as the label embedding matrix, where each column in $W_2$ corresponds to the embedding of a particular constituent label. $|H|$ represents the hidden dimension and $|L|$ is the size of the label set.

\subsection{Tree Scores and Chart-based Decoding}

The model assigns a score $s(T)$ to each tree $T$, which can be decomposed as:
\begin{equation}
s(T)=\sum_{(i,j,l) \in T}s(i,j,l)
\end{equation}


At test time, the model-optimal tree can be found efficiently using a CKY-style inference algorithm. Given the correct tree $T*$, the model is trained to satisfy the margin constraints:
\begin{equation}
s(T*)\geq s(T)+\Delta (T,T^{*})
\end{equation}
for all trees $T$ by minimizing the hinge loss:
\begin{equation}
max(0,\underset{T\neq T^{*}}{max }\left [s(T)+\Delta (T,T^{*}) \right ]-s(T^{*}))
\end{equation}
Here $\Delta$ is the Hamming loss on labeled spans, and the tree corresponding to the most-violated constraint can be found using a slight modification of the inference algorithm used at test time.

\begin{table*}[!tp]
	\centering 
	\begin{tabular}{l|c|c|c||c|c|c}
		\hline
		\multirow{2}*{\bf Method} & \multicolumn{3}{c||}{\bf Restaurant} & \multicolumn{3}{c}{\bf Laptop} \\ \cline{2-7}
		&   P. &   R. &   F1. &   P. &   R. &   F1. \\ \hline
		BERT-CRF & 0.3717 & 0.3055 & 0.3353 & 0.2966 & 0.2562 & 0.2749 \\ \hline 
		JET & 0.5731 & 0.2754 & 0.3720 & 0.4326 & 0.1435 & 0.2155    \\ \hline 
		TAS-BERT  & 0.2611 & 0.4509 & 0.3307 & \bf 0.4654 & 0.1892 & 0.2690  \\ \hline
		Extract-Classify & 0.3812 & 0.5144 & 0.4378 & 0.4523 & 0.2822 & 0.3475 \\ \hline \hline
		BARTABSA & 0.5793 & 0.5513 & 0.5650 & 0.4032 & 0.3853 & 0.3940  \\ \hline  
		GAS & 0.5871 & 0.5694 & 0.5781 & 0.3989 & 0.3917 & 0.3953 \\ \hline 
		Paraphrase & 0.5977 &  \bf0.6045 & 0.6011 & 0.3842 & \bf 0.3930 & 0.3885 \\ \hline 
		OTG & 0.6094 & 0.5988 & 0.6040 & 0.4102 &  0.3901 & 0.3998     \\ \hline \hline
		Ours & \bf 0.7113 & 0.5608 & \bf 0.6271 & 0.4512 & 0.3791 & \bf 0.4120     \\ \hline
	\end{tabular}
	\caption{Comparison with baselines.} \label{table-baseline}
\end{table*}

\section{Experiments}

In this section, we introduce the dataset used for evaluation and the baseline methods employed for comparison.
We then report the experimental results conducted from different perspectives. 

\subsection{Setting}
In this study, we use ACOS dataset~\cite{CaiXY20} for our experiments. 
There are 2,286 sentences in Restaurant domain, and 4,076 sentences in Laptop domain.
Following the setting from~\citet{CaiXY20}, we divide the original dataset into a training set, a validation set, and a testing set.
In particular, we remove some sentences (1.5\% among all the sentences) which cannot be parsed (e.g., one-to-many with implicit term, nested, overlapped). The distribution of the dataset can be found in Table~\ref{table-data}.

\begin{table}[!tp]
	\centering 
	\begin{tabular}{l|c|c|c}
		\hline
		\bf Domain & \bf Train & \bf Validation & \bf Test \\ \hline
		Restaurant & 1,529 & 171 &  582  \\ \hline
		Laptop & 2,929 &  326  &  816  \\ \hline
	\end{tabular}
	\caption{Distribution of the dataset.} \label{table-data}
\end{table}

We tune the parameters of our models by grid searching on the validation dataset.
For fair comparison, we employ T5~\cite{Raffel2020T5}  and fine-tune its parameters not only for our opinion tree parser’s encoder, but also for the backbone of all other generative methods. The model parameters are optimized by Adam~\cite{kingma2014adam} with a learning rate of 5e-5. The batch size is 128 with a maximum 512 token length. Our experiments are carried out with a Nvidia RTX 3090 GPU.
The experimental results are obtained by averaging ten runs with random initialization.

In evaluation, a quadruple is viewed as correct if and only if the four elements, as well as their combination, are exactly the same as those in the gold quadruple. 
On this basis, we calculate the Precision and Recall, and use F1 score as the final evaluation metric for aspect sentiment quadruple extraction~\cite{CaiXY20,ZhangD0YBL21}.

\subsection{Main Results}
We compare the proposed opinion tree parser with several classification-based aspect-based sentiment analysis models, including,  \emph{BERT-CRF}~\cite{bert18}, \emph{JET}~\cite{XuLLB20}, \emph{TAS-BERT}~\cite{WanYDLQP20} and \emph{Extract-Classify}~\cite{CaiXY20}.
In addition, generative models are also compared, such as \emph{BARTABSA}~\cite{YanDJQ020}, \emph{GAS}~\cite{Zhang0DBL20}, \emph{Paraphrase}~\cite{ZhangD0YBL21} and \emph{OTG}~\cite{BaoWJXL22}.\footnote{ The implementations of  JET, TAS-BERT, Extract-Classify and OTG are based on their official codes, we re-implement the remaining by ourselves.}

As shown in Table~\ref{table-baseline}, we find that generative models give the best performance among the previous systems. It shows that the unified generation architecture helps extract sentiment elements jointly.
Meanwhile, our proposed model outperforms all the previous studies significantly ($p<0.05$) in all settings. 
It indicates that the chart-based opinion parser is more useful for explicitly modeling tree structural constraints, while previous generative models cannot guarantee the structure well-formedness, and their generated linearized string ignores the implicit alignments among sentiment elements.
Furthermore, the results also indicate the effectiveness of the context-free opinion grammar, which is used to form the sentiment structure into an  opinion tree.

\subsection{Comparison of Decoding Efficiency}

\begin{table}[!tp]
	\centering 
	\begin{tabular}{l|c|c}
		\hline
		\bf Method & \bf Encoder &  \bf Time (s) \\ \hline
            BERT-CRF  &  \multirow{3}{*}{BERT}  &  1.96  \\ \cline{1-1} \cline{3-3}
            JET &  &  2.83  \\ \cline{1-1} \cline{3-3}
            Ours &  &  \bf 0.81 \\ \hline \hline
		GAS & \multirow{4}{*}{T5}  &  58.2  \\ \cline{1-1} \cline{3-3}
		Paraphrase &  &  61.3  \\ \cline{1-1} \cline{3-3}
		OTG  &  & 64.9   \\ \cline{1-1} \cline{3-3} 
		Ours   &   & \bf1.04   \\ \hline 
	\end{tabular}
	\caption{Decoding efficiency of different models.} \label{table-time}
\end{table}

Table~\ref{table-time} compares different models in terms of decoding speed. 
For a fair comparison, we re-run all previous models on the same GPU environment. 
The results are averaged over 3 runs. In addition, the settings of batch size are the same for all the models. 

As we can see, for generative models~\cite{Zhang0DBL20,ZhangD0YBL21,BaoWJXL22}, they have to generate words one by one, leading to their low speed, and the beam searching during decoding makes the speed much slower.
Meanwhile, based on span-based searching, our chart-based opinion tree parser achieves a much higher speed. 
In addition, the speed of proposed opinion tree parser is faster than the classification-based models (e.g., BERT-CRF, JET). It may be due to that these classification-based models extract the sentiment elements one by one as pipeline systems.
It also indicates the effectiveness of the chart-based parser and span-based searching, which could parallelly extract the sentiment elements in the sentence.

\section{Analysis and Discussion}

In this section, we give some analysis and discussion to show the effectiveness of proposed opinion tree parser for aspect-based sentiment analysis.

\subsection{Effect of Context-Free Opinion Grammar}


\begin{figure}[t]
	\centering
	\includegraphics[width=225px]{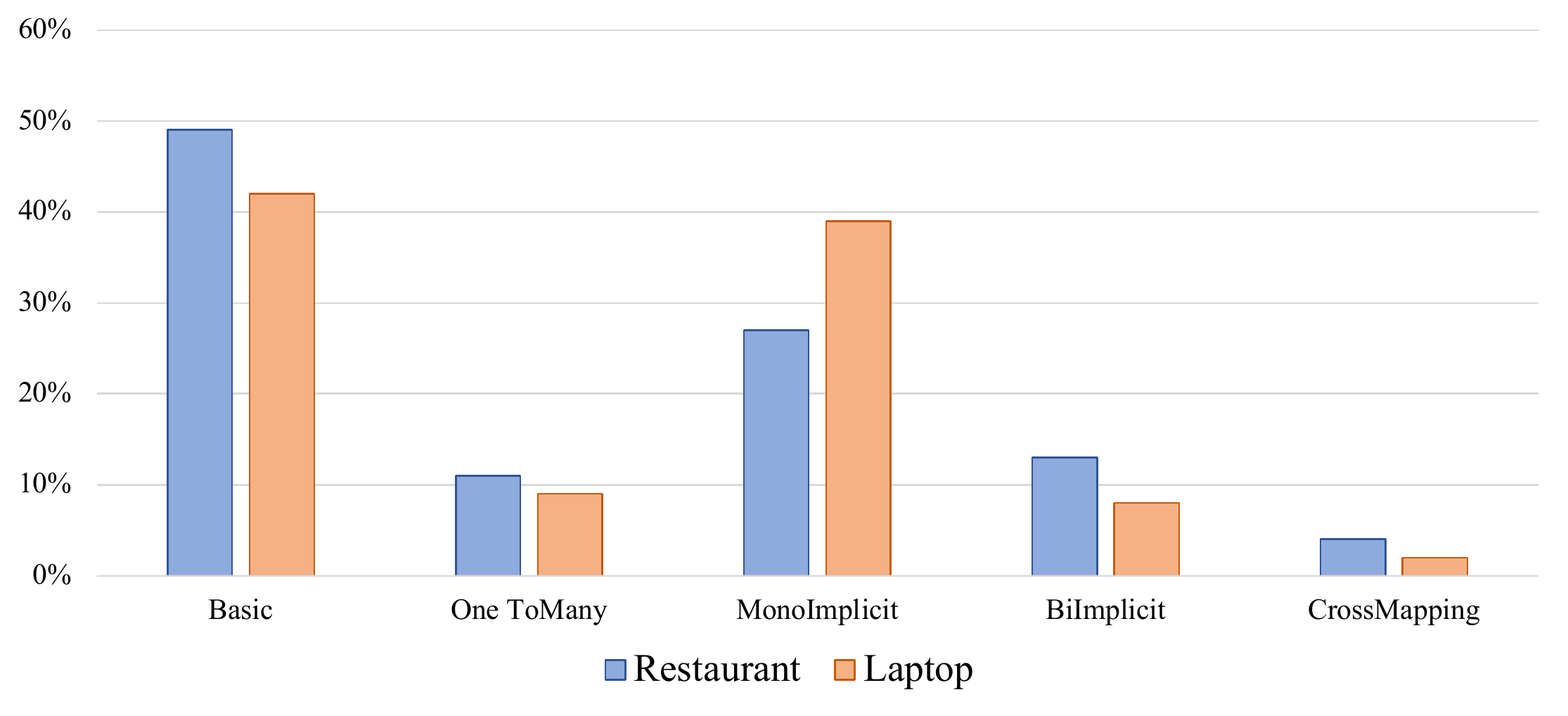}
	\caption{\label{figure-percent} Statistic of  regular and irregular situations of opinion trees.}
\end{figure}

\begin{table}[!tp]
	\centering 
	\begin{tabular}{l|c|c}
		\hline
		\bf Rules & \bf Restaurant & \bf Laptop \\ \hline
		Basic & 0.4558 & 0.2727     \\ \hline \hline
		\ +OneToMany & 0.5812 & 0.3175     \\ \hline
		\ +MonoImplicit & 0.4856 &  0.3632    \\ \hline
		\ +BiImplicit & 0.5167 &  0.2984    \\ \hline
		\ +CrossMapping & 0.4598 &  0.2786    \\ \hline \hline
		Ours & 0.6271 & 0.4120     \\ \hline
	\end{tabular}
	\caption{Results of different conditional rules in context-free opinion grammar  with F1-score measurement.} \label{table-rule}
\end{table}

We firstly give the statistic of regular and irregular situations of opinion trees in Figure~\ref{figure-percent}, where $Basic$ is the regular situation which contains full four elements of a quadruple, and others are the irregular situations. From the figure, we find that the distribution of these situations are similar in the two domains: around half of reviews contains regular full quadruple situations, and mono-implicit is the most frequency irregular situations.

We then analyze the effect of different conditional rules which are used to solve irregular situations.  As shown in Table~\ref{table-rule}, we can find that if we only use the basic rules, the performance of opinion tree parser is very low.
It may be due to the irregular situations appear in more than half of the review texts. 
In addition, all the conditional rules are beneficial to parse the opinion tree. Among these rules, one-to-many performs better than others.
Furthermore, our proposed model achieves the best performance, which proves the effect of conditional rules.

\subsection{Results of Different Tree Parsers}


\begin{table}[!tp]
	\centering 
	\begin{tabular}{l|c|c}
		\hline
		\bf Method  & \bf Restaurant & \bf Laptop \\ \hline
        BERT-CRF  & 0.3353 & 0.2749     \\ \hline \hline
		Zhang19  & 0.5021 & 0.3537     \\ \hline
		Nguyen21 & 0.5872 & 0.3673    \\ \hline
        Yang22  & 0.5936 & 0.3712    \\ \hline \hline
		Ours  & 0.6123 & 0.3748    \\ \hline
	\end{tabular}
	\caption{Results of different parsers with F1-score measurement.} \label{table-parser}
\end{table}

We then analyze the effect of different tree parsers with the proposed context-free opinion tree grammar. In particular, we select three popular  parsers which have shown their effect on syntax tree parsing~\cite{Junchi19,Nguyen21} and name entity recognition~\cite{Yangtu22}. Among these parsers, \citet{Junchi19} is transition-based parser, which constructs a complex output structure holistically, through a state-transition process with incremental
output-building actions; \citet{Nguyen21} and \citet{Yangtu22}  are sequence-to-sequence parsers, which employ pointing mechanism for bottom-up parsing and use sequence-to-sequence backbone.
For fair comparison, we use RoBERTa-base~\cite{roberta} as the backbone of all the parsers and our proposed chart-based opinion tree parser.


As shown in Table~\ref{table-parser}, all the parsers outperform the BERT-CRF. It shows the effect of the proposed context-free opinion grammar. 
No matter which parser we use, it achieves better performance than classification-based models.
In addition, our chart-based opinion tree parser outperforms all the other parsers with a remarkable advantage. 
It may be due to that all the other parsers suffer from error propagation and exposure bias problems. 
Meanwhile, our proposed chart-based parser could infer parallelly, especially effective in parsing long review texts. 
Such observation has also been proven in neural constituency parsing~\cite{CuiY022}, the chart-based parser reported state-of-the-art performance in that task.


\subsection{Impact of Opinion Tree Schemas}

\begin{table}[!tp]
	\centering 
	\begin{tabular}{c|c|c|c}
		\hline
		\bf Schema & \bf Domain & \bf OTG & \bf Ours \\ \hline
		\multirow{2}{*}{Pair} & Restaurant & 0.6906 &   0.7681    \\ \cline{2-4}
		 & Laptop & 0.7201 &   0.7602    \\ \hline
		\multirow{2}{*}{Triple} & Restaurant  & 0.6582 &   0.7051    \\ \cline{2-4}
		 & Laptop   & 0.6562 &   0.6843    \\ \hline 
		\multirow{2}{*}{Quad}  & Restaurant &   0.6040 & 0.6271 \\ \cline{2-4}
		 & Laptop &   0.3998 & 0.4120     \\ \hline
	\end{tabular}
	\caption{Results on different opinion tree schemas with F1-score measurement.} \label{table-tree}
\end{table}

We analyze the effect of the proposed model with the opinion tree generation model (OTG)~\cite{BaoWJXL22} in different opinion tree schemas.
OTG employs a generative model to jointly detect all sentiment elements in a linearized tree formation with a sequence-to-sequence architecture.
In particular, there are three popular schemas: \emph{Pair} means that we only extract aspect term and opinion term from review text~\cite{QiuLBC11,XuLLB20,li-etal-2022-generative}, and \emph{Triple} means that we extract aspect term, opinion term, and polarity from review text~\cite{Zhang0DBL20,ChenWLW21}. \emph{Quad} is the quadruple schema that extracts the whole four sentiment elements to form the opinion tree~\cite{CaiTZYX20,ZhangD0YBL21,BaoWJXL22}.
Note that,  we make minor modifications to the context-free opinion grammar, and let it suitable for Pair and Triple schemas.

From Table~\ref{table-tree}, we can find that our model outperforms OTG in all the schemas. It indicates that our opinion tree parser model is generalized and can be used to handle different schemas in aspect-based sentiment analysis.
It also shows that the parsing strategy is more effect than generative model on capture the structure of sentiment elements.
In addition, we also find that the improvement of Pair and Triple are much higher than Quad, it may be due to that the simple schema is easier to normalize and recover.

\begin{figure}[t]
	\centering
	\includegraphics[width=230px]{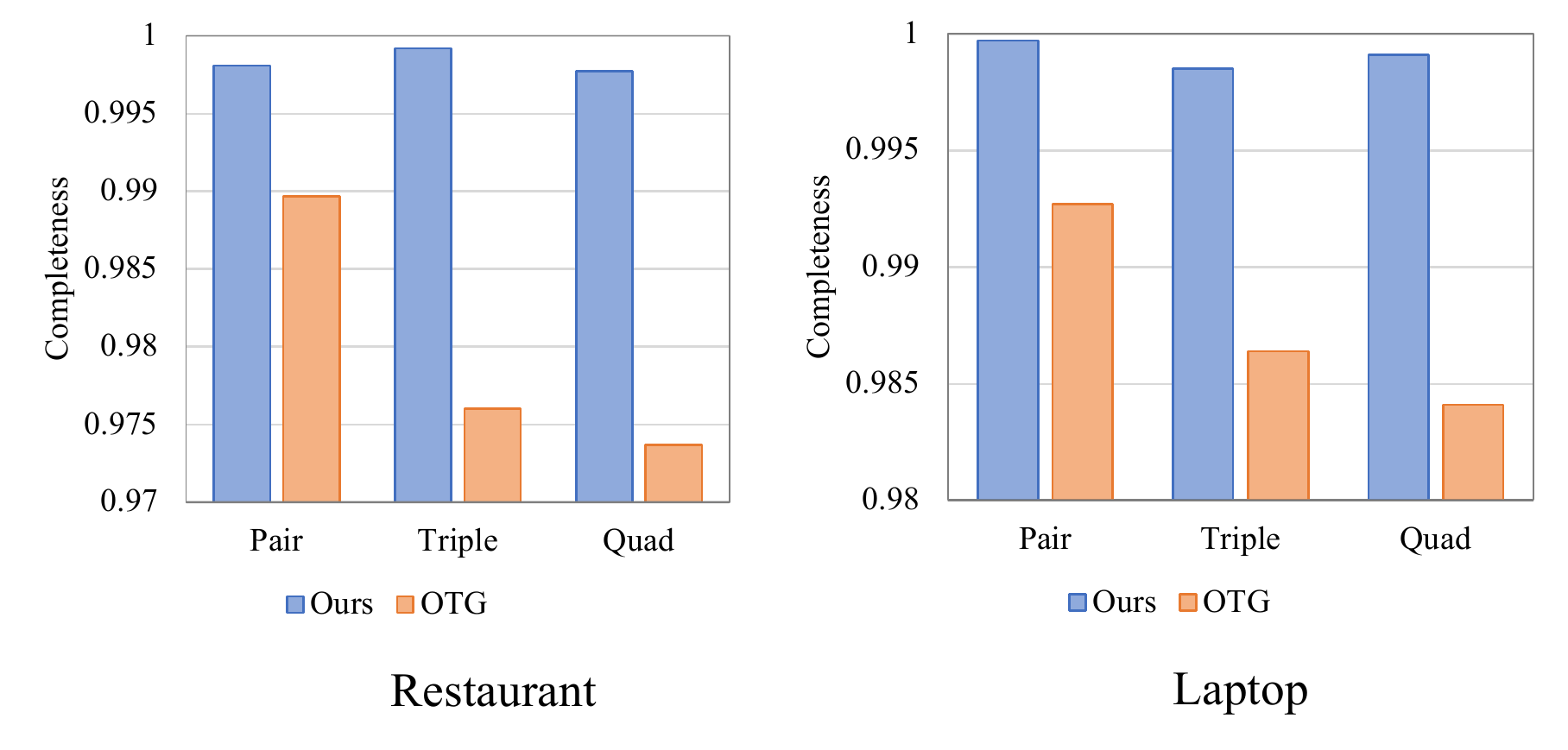}
	\caption{\label{figure-completeness} Tree structure completeness of different models.}
\end{figure}
 
We then analyze the completeness of the tree structure generated/parsed from OTG and the proposed model. 
The completeness is calculated through the valid rate of a tree structure.
As shown in Figure~\ref{figure-completeness}, the completeness of the proposed model is higher than OTG in all the schemas.
It shows that our proposed model can explicitly model tree structural constraints, and guarantee tree structure well-formedness.
In addition, the high completeness also guarantees the quality of recovery from tree structure to sentiment elements. 

Furthermore, case studies in Appendix~\ref{appendixB} are given to make more intuitive comparisons between OTG and proposed opinion tree parser.

\section{Conclusion}

In this study, we propose a novel opinion tree parsing model, aiming to parse all the sentiment elements into an opinion tree, which  can reveal a more comprehensive and complete aspect-level sentiment structure.
In particular, we first introduce a novel context-free opinion grammar to normalize the opinion structure. We then employ a neural chart-based opinion tree parser to fully explore the correlations among sentiment elements and parse them in the opinion tree form.
Detailed evaluation shows that our model significantly advances the state-of-the-art performance on several benchmarks. 
The empirical studies also show that the proposed opinion tree parser with context-free opinion grammar is more effective in capturing the opinion tree structure than generative models with a remarkable advantage in computation cost. 

\section{Limitations}

% ...

The limitations of our work can be stated from two perspectives. First, the proposed context-free opinion grammar is designed manually. It can be the future work to explore how to automatic generate the grammar.
Secondly, we focus on opinion tree parsing in one major language. The performance of other languages remains unknown.

\section*{Acknowledgments}

We would like to thank Prof. Yue Zhang for his helpful advice and discussion during this work.
Also, we would like to thank the anonymous reviewers for their excellent feedback.
This work is supported by the China National Key R\&D Program (No. 2020AAA0108604), and the National Natural Science Foundation of China (No. 61976180, No. 62006093).


\bibliography{anthology,custom}
\bibliographystyle{acl_natbib}

\appendix

\section{Tree Pruning}\label{appendixA}

\begin{figure*}[t]
	\centering
	\includegraphics[width=350px]{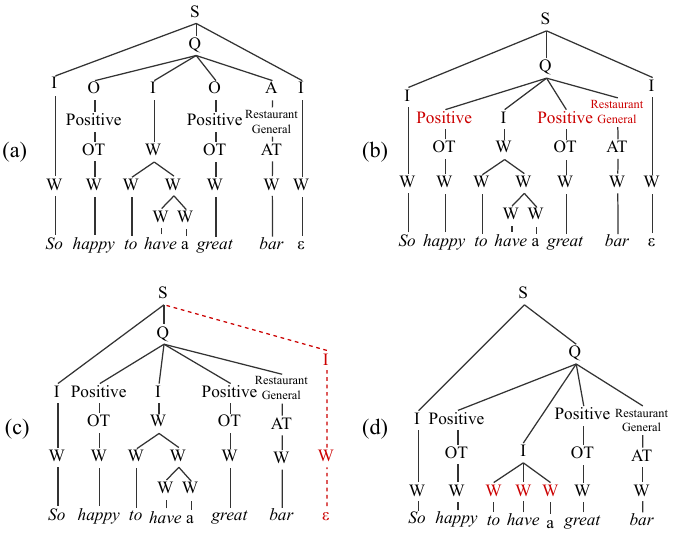}
	\caption{\label{original_parsing_tree} Example of tree pruning.}
\end{figure*}

As the original opinion trees are too complicated for parsing, we adopt a pruning method to reduce the duplication in trees. To be more specific, we introduce our method with a pruning example of review ``So \emph{happy} to have a \emph{great} bar'', which can be described as following steps, and the original tree is demonstrated in  Figure~\ref{original_parsing_tree}(a).
\begin{itemize}

\item The unary chain of category and polarity are integrated into the aspect node and opinion node respectively.
The processed result is shown in Figure~\ref{original_parsing_tree}(b).

\item We delete the chains with $\epsilon$ leaf node, the processed result is shown in Figure~\ref{original_parsing_tree}(c).

\item If the  children nodes contain nodes that have exactly the same node type with the parent node, we will delete the parent node and connect children with the ancestor node directly, the processed result is shown in Figure~\ref{original_parsing_tree}(d).

\end{itemize}

Therefore, Figure~\ref{original_parsing_tree}(d) gives the final formation of our opinion tree for parsing.

\section{Case Study}\label{appendixB}

We launch a set of case studies to make a more intuitive comparison between our model and OTG~\cite{BaoWJXL22}.
We select reviews that are predicted into invalid formation by OTG to demonstrate our models' superiority in guaranteeing structure well-formedness. 
As demonstrated in  Table~\ref{case_study}, these cases can be divided into following categories:

\subsection*{Invalid Term}

The first three examples are about invalid terms which generated from OTG. 

In the first example, OTG gives a very typical wrong prediction, it rewrites ``\emph{waiting}'' to "\emph{wait}", which could change the original meanings and does not meet the requirement of extracting raw text from the review, while our method operating over raw spans, easily gives a right answer.
 
In the second example, OTG generates ``\emph{atmosphere}'' as the aspect term based on its understanding of ``\emph{feeling}'' since they have similar semantic information.
However, `\emph{atmosphere}'' does not exist in the review. 
On the other hand, our model also shots the right target but selects it as the final prediction under the constraints of chart decoder.

In the third example, OTG generates ``\emph{not that slow}'' from the review, which are not continuous in the original text: the words ``\emph{not that}'' appear in the beginning but ``\emph{slow}'' appears in the end. 
In this situation, our span-based method can easily extract "\emph{slow}" as the opinion term since it can only operate over raw spans.

\subsection*{Invalid Structure} 

The invalid structure means that the output sequence of OTG can not be recovered into a valid tree structure, this may due to various reasons. One of the common reasons is unmatched brackets.
The fourth example shows an OTG's output sequence that can not be decoded into a valid tree since the sequence that starts with  ``opinion'' can not be recognized as a subtree. 
In contrast, with the CYK-style algorithm, our method build trees and subtrees over spans, ensuring the legality of trees or subtrees.

\subsection*{Invalid Category} 

OTG also would classifies  aspect term into a non-existing category. In the fifth example, the aspect term ``\emph{msi headset}'' is classified into a non-existing category "HEADSET GENERAL" by OTG, which usually happens when it comes to the generative method with LAPTOP dataset since it has more than 100 categories. 
This would not be a difficult problem for our model's classifier, it will set specific target classes before starting the training process.\newline

\begin{table*}[!tp]
	\centering 
	\begin{tabular}{l|c|c|c}
		\hline
		\bf Review text & \bf Reason & \bf OTG & \bf Ours  \\ \hline
  
	   \makecell[l] {The waiting staff \\ has been perfect} & \makecell[c] {Invalid \\ Term} &
                \makecell[l]{SERVICE GENERAL\textcolor{blue}{\checkmark} \\ wait staff \textcolor{red}{\ding{55}} \\POSITIVE \textcolor{blue}{\checkmark} \\ perfect \textcolor{blue}{\checkmark}}& 
                \makecell[l]{SERVICE GENERAL \textcolor{blue}{\checkmark} \\ waiting staff \textcolor{blue}{\checkmark}\\POSITIVE \textcolor{blue}{\checkmark} \\ perfect \textcolor{blue}{\checkmark}}     \\ \hline
		 
          \makecell[l] {I also really enjoy\\ the intimate feeling of\\ a small restaurant.} & \makecell[c] {Invalid \\ Term}  &
                \makecell[l]{AMBIENCE GENERAL\textcolor{blue}{\checkmark} \\ atmosphere\textcolor{red}{\ding{55}}\\ POSITIVE \textcolor{blue}{\checkmark} \\ intimate \textcolor{blue}{\checkmark}} &
                \makecell[l]{AMBIENCE GENERAL \textcolor{blue}{\checkmark} \\ feeling \textcolor{blue}{\checkmark}\\ POSITIVE \textcolor{blue}{\checkmark} \\ intimate \textcolor{blue}{\checkmark}}     \\ \hline
          
          \makecell[l] {not that this machine \\ boots up slow.} & \makecell[c] {Invalid \\ Term}  &
                \makecell[l]{OS PERFORMANCE\textcolor{red}{\ding{55}} \\ boots up\textcolor{blue}{\checkmark}\\ neural \textcolor{red}{\ding{55}} \\ not that slow \textcolor{red}{\ding{55}}} &
                \makecell[l]{LAPTOP PERFORMANCE \textcolor{blue}{\checkmark} \\ boots up \textcolor{blue}{\checkmark}\\ NEGATIVE \textcolor{blue}{\checkmark} \\ slow \textcolor{blue}{\checkmark}}     \\ \hline
                
          \makecell[l] {we're can't say enough \\ about  their delicious\\ gourmet pizza ’ s!} &  \makecell[c] {Invalid \\ structure} &
                \makecell[l]{( root  ( quad ( aspect \\( food quality, pizza ) ) ), \\ ( opinion ( positive, null ) ) ) )\textcolor{red}{\ding{55}}} &
                \makecell[l]{FOOD QUALITY\textcolor{blue}{\checkmark} \\ delicious \textcolor{blue}{\checkmark}\\ POSITIVE \textcolor{blue}{\checkmark} \\ pizza ’ s \textcolor{blue}{\checkmark}}     \\ \hline
        
          \makecell[l] {writing this review so \\ early to receive that\\ nice msi headset.} &  \makecell[c] {Invalid \\ category} &
                \makecell[l]{HEADSET GENERAL\textcolor{red}{\ding{55}} \\ msi headset\textcolor{blue}{\checkmark}\\ POSITIVE \textcolor{blue}{\checkmark} \\ nice \textcolor{blue}{\checkmark}} &
                \makecell[l]{DEVICE GENERAL \textcolor{blue}{\checkmark} \\ msi headset\textcolor{blue}{\checkmark}\\ POSITIVE \textcolor{blue}{\checkmark} \\ nice \textcolor{blue}{\checkmark}}    \\ \hline
	\end{tabular}
	\caption{Case study} \label{case_study}
\end{table*}
From the cases shown in Table~\ref{case_study}, we can find that our method shows significant superiority in modeling tree structural constraints and guaranteeing tree structure well-formedness, along with the quality of recovery from tree structure to sentiment elements, while OTG has to employ complex post-processing method to strengthen its shortage.


\end{document}